\pdfoutput=1
\documentclass[11pt]{article}

\usepackage[]{EMNLP2023}

\usepackage{times}
\usepackage{latexsym}
\usepackage{graphicx}
\usepackage{booktabs}
\usepackage{multirow}
\usepackage{tabu}

\usepackage{array}
\newcommand{\PreserveBackslash}[1]{\let\temp=\\#1\let\\=\temp}
\newcolumntype{C}[1]{>{\PreserveBackslash\centering}p{#1}}
\newcolumntype{R}[1]{>{\PreserveBackslash\raggedleft}p{#1}}
\newcolumntype{L}[1]{>{\PreserveBackslash\raggedright}p{#1}}

\usepackage{algorithm}
\usepackage{algpseudocode}

\usepackage[T1]{fontenc}
\usepackage[utf8]{inputenc}

\usepackage{microtype}

\usepackage{inconsolata}

\title{A surprisal oracle for when every layer counts}

\author{
  Xudong Hong\footnotemark[3]\ ,  Sharid Loáiciga\footnotemark[2] \ and Asad Sayeed\footnotemark[2] \ \\[0.8ex]
  \footnotemark[3] Dept. of Language Science and Technology and Dept. of Computer Science,
  Saarland University \\ 
  \footnotemark[2] Dept. of Philosophy, Linguistics, and Theory of Science, University of Gothenburg \\
  {\tt \{xhong\}@lst.uni-saarland.de}, 
  {\tt \{sharid.loaiciga,\ asad.sayeed\}@gu.se}\\
}

\begin{document}
\maketitle
% \blfootnote{$\ast$ These authors contributed equally to this work.}

\begin{abstract}
%We investigate the viability of surprisal in an active curriculum learning framework to train transformer-based language models in the context of the BabyLM Challenge. In our approach, the model itself selects the data to label (active learning) and schedules data samples based on a surprisal oracle (curriculum learning). We show that the models learn across all the tasks and datasets evaluated, making the technique a promising alternative approach to reducing the data requirements of language models. 
Active Curriculum Language Modeling \citep[ACLM;][]{hong-etal-2023-surprisal} is a learner-directed approach to training a language model.  We proposed the original version of this process in our submission to the BabyLM 2023 task, and now we propose an updated ACLM process for the BabyLM 2024 task. ACLM involves an iteratively- and dynamically-constructed curriculum informed over the training process by a model of uncertainty; other training items that are similarly uncertain to a least certain candidate item are prioritized. Our new process improves the similarity model so that it is more dynamic, and we run ACLM over the most successful model from the BabyLM 2023 task: ELC-BERT \citep{charpentier-samuel-2023-layers}. We find that while our models underperform on fine-grained grammatical inferences, they outperform the BabyLM 2024 official baselines on common-sense and world-knowledge tasks.  We make our code available at \url{https://github.com/asayeed/ActiveBaby}.
\end{abstract}

\section{Introduction}

In this work, we describe our contribution to the "strict-small" task of the BabyLM Challenge of 2024 \citep{warstadt2024call} which follows up our contribution to BabyLM 2023 \citep{hong-etal-2023-surprisal}. Our effort this year focused on two activities: (1) testing the most successful contribution to BabyLM 2023, Every Layer Counts BERT \citep[ELC-BERT;][]{charpentier-samuel-2023-layers} under additional conditions and (2) implementing our training protocol, which we called Active Curriculum Language Modeling (ACLM) over our attempt at replicating ELC-BERT.  We test ELC-BERT under more constrained conditions and explore whether the result is stable under other hyperparameter settings.  Under very similar settings, we test our ACLM approach to see whether it exceeds the performance of our baselines on the BabyLM evaluation tasks.

Our intuition is that a human learner is an active participant in the environment of language acquisition \citep{fazekas2020do,masek2021where}.  That children are not passive participants in L1 acquistion goes essentially without saying in contemporary developmental psycholinguistics---and with anyone who has interacted with a small child for any length of time---but the best artificial learners under data-constrained conditions (such as ELC-BERT) are trained in an entirely passive way.  Their higher performance stems entirely from technical adjustments to the "training math".  While this results in impressive performance, the insights it can give to the "whole picture" of how children acquire language from small data is limited, given what we know already about human development.

\begin{figure*}[ht!]
    \centering
    \includegraphics[width=0.99\linewidth, trim=0cm 11.5cm 5cm 0cm, clip=TRUE]{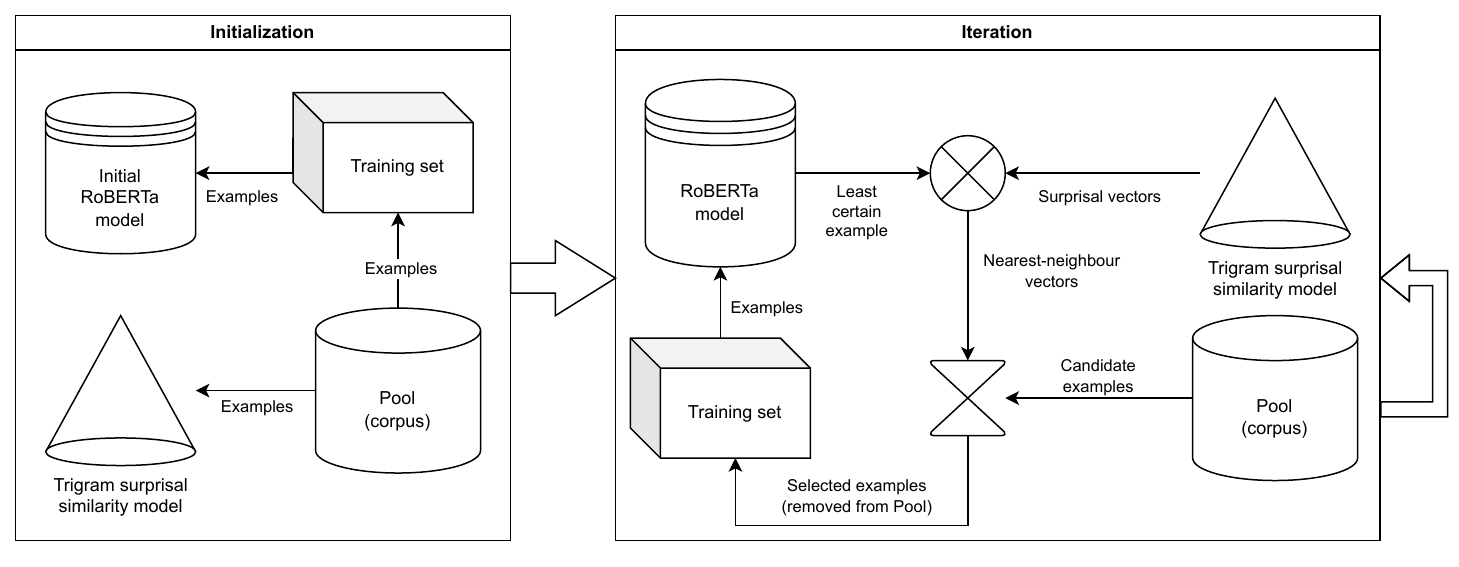}
    \caption{The architecture of our ACLM method from last year's submission, described in \citet{hong-etal-2023-surprisal}.  For this study, we modify the trigram surpisal similarity model to simply use the average sentence surprisal of the model under training, which is now ELC-BERT rather than RoBERTa.}
    \label{fig:model}
\end{figure*}

Instead, our over-arching hypothesis is that for every successful "passive" language modeling training algorithm, there is a way of scheduling the learning process that is more cognitively plausible or better-performing or both.  This is not straightforwardly "classical" curriculum learning with the curriculum calculated or set in advance.  Rather, it takes its inspiration from active learning \citep{zhang-etal-2022-survey}, where the learner (usually in a classification task) assesses its uncertainty on hitherto unseen items, and then asks for a human label, in a process that reduces the burden of labelling more training data than there are resources to label.

ACLM instead uses a cycle in which the learner trains an initial model from a small subset of the examples, and then iteratively adds to its dataset by using an uncertainty criterion over the items automatically, essentially creating a "dynamic" curriculum during the learning process \citep{bengio_2009_curriculum,jafarpour-etal-2021-active}.  

The outcome of the overall BabyLM 2023 task participation \citep{warstadt2023papers} suggested that curriculum learning was not fruitful in exceeding the original baselines or in overall competitiveness on the BabyLM task as compared to model architecture "tweaks" such as ELC-BERT.  The present study suggests a more mixed picture: that the advantages from architectural modifications are highly sensitive to peturbations from hyperparameters, while a dynamically updated curriculum such as ACLM still may have the potential to augment a high-performing model architecture while retaining some connection to the interactive nature of human language acquisition.

\section{Background}
The BabyLM task necessarily involves the exploration of a very large solution space. 
In the previous year's challenge, we proposed an initial system, which we depict in fig. \ref{fig:model}.  Because we had to start somewhere, this involved design decisions based on educated guesses as well as a focus on efficiency and "getting it off the ground".  The result of that effort was that our system ended in the "middle of the pack" and behind a baseline BERT model in the actual competition, but nevertheless resulted in insights that led us to consider how we can continue to explore this part of the BabyLM solution space, considering our expectations from a cognitive perspective.  

Our previous system started with a blank RoBERTa model \citep{liu2019roberta,zhuang-etal-2021-robustly} which was initialized by training on a small subset of the training corpus.  The remainder of the training corpus (the "pool") was processed via a trigram model into sequences of per-token trigram surprisal values, where suprisal is defined as the negative log-probability of the trigram ending in the given token.  These sequences were resampled into an arbitrarily-chosen seven dimensions (which we later found was the average token length of the samples in the corpus), which we call the "surprisal signature".

At each iteration of training (several epochs), the RoBERTa model is queried about every sentence in the current training set, starting from the initial subset: which previously-seen utterance had the highest average surprisal.  Then a k-nearest-neighbours process was used to sample the most similar surprisal signatures to the signature of the least certain sentence.  These are added to the active training set, and the next iteration commences.   

This process is different from the active curriculum learning process of \citet{jafarpour-etal-2021-active}. \citeauthor{jafarpour-etal-2021-active} develop a way to combine a human expert-designed curriculum with an active-learning informativeness criterion in order to select instances for humans to label. Since our training task is a language modeling task rather than a labeling or classification task, we can eliminate the human in the loop (effectively, the label is a word we already have in the text) and use the informativeness criterion to structure an automatic curriculum.

One obvious weakness of this process is that the surprisal space is static and does not reflect changes in the learner's estimation of what is surprising with respect to what is being learned.  There are other weaknesses of this process, such as the seeming arbitrariness of the seven-dimensional vectors or even the use of surprisal as the criterion itself.  
However, in this update of our previous work, besides replacing RoBERTa with ELC-BERT, we focus on the first weakness and implement a dynamic process as described below.

\section{Learner-directed Active Curriculum Language Modeling}
In the original formulation of ACLM, the "surprisal space" (the collection of "surprisal signature" vectors), was static throughout training, leading to a curriculum directed entirely by the model's uncertainty over each active learning iteration. From a cognitive perspective, this would be equivalent to a human learner whose expectations about the most educational thing in the environment never change from birth.  We now propose an update to ACLM: we generate a new suprisal space at every iteration, which more closely matches the idea that a learner changes its view of the learning environment as it learns. In practical terms, this means that our ACLM model now re-evaluates the suprisal space using the ELC-BERT model itself, producing a new surprisal space reflecting the model's current "knowledge state". We view this as increasing the cognitive realism of ACLM and making it reflect a more learner-directed approach to acquisition.  

We describe the ACLM procedure in this year's submission at a high level. We refer to "iterations" of the training procedure to be periods between updates of the active training set from the pool (corpus of as-yet-unseen training items). Multiple training epochs can take place during an iteration, meaning that the model may see the same training set without an update multiple times. Algorithms \ref{alg:init} and \ref{alg:iter} provide an overview of the process, with the former describing the initialization process and the latter the iterative curriculum adaptation.  The biggest procedural difference between this and \citet{hong-etal-2023-surprisal} is the use of the model itself to update the surprisals at every iteration. 
\begin{algorithm}[t]
\begin{algorithmic}
    \State Model $\gets$ new(ELC-BERT)
    \State ActiveSet $\gets$ select\_random(Pool, $n\_$\textit{initial})
    \State train(Model, ActiveSet, $n\_$\textit{epochs})
    \State SurprisalSet $\gets$ []
    \ForAll{instances $i$ in Pool}
        \State \textit{surprisals} $\gets$ Model.surprisals($i$)
        \State SurprisalSet.append(\textit{surprisals})
    \EndFor
\end{algorithmic}
\caption{Initialization phase of this year's ACLM process.}\label{alg:init}
\end{algorithm}

\begin{algorithm}[t]
\begin{algorithmic}
    \For{\textit{iter} $\gets$ 0 to $n$\_\textit{iterations}}
        \State \textit{max\_surprised} $\gets$ TrainingSet[0]
        \ForAll{instances $i$ in TrainingSet}
            \State \textit{orig\_surprisal} $\gets$ \\\hfil \hfil Model.surprisals(\textit{max\_surprised}) 
            \State \textit{new\_surprisal} $\gets$ Model.surprisals($i$) 
            \If{\textit{orig\_surprisal} $<$ \textit{new\_surprisal}}
                \State \textit{max\_surprised} $\gets$ $i$
            \EndIf
        \EndFor
    \State ActiveSet.update(SurprisalSet.kNN(\\\hfil\textit{max\_surprised}, $k$, Pool))
    \State train(Model, ActiveSet, $n\_$\textit{epochs})
    \State SurprisalSet $\gets$ []
    \ForAll{instances $i$ in Pool}
        \State \textit{surprisals} $\gets$ Model.surprisals($i$)
        \State SurprisalSet.append(\textit{surprisals})
    \EndFor
    \EndFor
\end{algorithmic}
\caption{Iterations of the ACLM process. The kNN procedure also removes the instances from the Pool.}\label{alg:iter}
\end{algorithm}

In our previous submission, we split the corpus by utterance.  This year, we follow the practice of the ELC-BERT implementation of having a sequence length of 128 tokens regardless of utterance boundaries. In the surprisal space, our dimensionality reduction proceeds to 7 dimensions (D7, as in our previous submission), 32 dimensions (D32), 64 dimensions (D64), and 128 dimensions (D128, essentially with no reduction).  The reduction of the surprisal space no longer represents an attempt to equalize sentences of varying lengths through image resampling\footnote{We simply use the \texttt{resize} method from scikit-image.}, since everything starts from 128 tokens.

\section{Analysis}
We list our results in table \ref{tab:bblm2024}. The LTG-BERT baseline for the strict-small BabyLM 2024 task was trained with a batch size of 32786 and 8196 as well as corresponding sequence lengths of 512 and 128 \citep{samuel-etal-2023-trained}. The equivalent ELC-BERT run for BabyLM 2023 was also trained with a batch size of 8096 and a sequence length of 128 \citep{charpentier-samuel-2023-layers}. This is very resource-intensive, so we instead trained non-ACLM models with batch sizes of 32 and 512 (ELC-BERT B32 and ELC-BERT B512). Gradient accumulation was used as well to mitigate the smaller batch sizes (complete list of hyperparameters in Table \ref{tab:hyperparameters}, section \ref{sec:appendix}).  

The original ELC-BERT still vastly outperforms both the BabyLM 2024 strict-small baselines as well as all of our models on BLiMP and GLUE.  We will not attempt to explain ourselves why the 2024 baselines underperform the 2023 ELC-BERT submission\footnote{One reviewer suggests that this may partly be the result of a switch in averaging procedure in the evaluation pipeline provided by the task organizers.} and focus our discussion on this year's baselines.

On BLiMP \citep{warstadt2020blimp}, which contains inferences over very fine-grained grammatical details (e.g., anaphor agreement, island phenomena), our non-ACLM models do relatively poorly compared to LTG-BERT and BabyLlama baselines.  As the main difference is batch size, it is hard to speculate on any deeper reason for the lower performance.  This is essentially a candidate for an "unprincipled" hyperparameter search, as it is hard to imagine what batch size specifically has to do with grammatical phenomena.  Our ACLM models outperform our non-ACLM models slightly, but which ACLM models do best is not consistent over the supplement or the filtered portion of BLiMP. However, the overall consistency of outperformance of ACLM on the filtered BLiMP suggests that ACLM is having an effect.

On EWOK \citep{ivanova2024elements}, which is a dataset of inferences over world knowledge, we have a completely different story.  Our small batch-size non-ACLM ELC-BERT does far better than either BabyLlama or LTG-BERT. Our ACLM runs do even better than our non-ACLM ELC-BERT runs.  There is no strong difference between any degree of dimensionality reduction for the surprisal space.

On GLUE \citep{wang2018glue}, our non-ACLM ELC-BERT models are in the same range as the LTG-BERT and BabyLlama baselines.  However, our ACLM runs are all superior to any model but the original ELC-BERT. We do not see any potential for speculation on the performance differences for the surprisal space dimension.

\begin{table*}[th!]
    \centering
    %\resizebox{\textwidth}{!}{% 
\begin{tabular}{lcccc}
\toprule
\textbf{Model}  & \textbf{BLiMP suppl.} & \textbf{BLiMP filtered} & \textbf{EWOK} & \textbf{GLUE} \\
\midrule
ELC-BERT (original) & 67.9 & 80.5 & - & 75.3 \\
\midrule
BabyLlama & 59.5 & 69.8 & 50.7 & 63.3\\
LTG-BERT &60.8 & 60.6 & 48.9 & 60.3 \\
\midrule
ELC-BERT B32 & 50.1&47.9 & 65.2 & 63.4 \\
ELC-BERT B512 &47.8&49.1 & 64.9 & 61.0 \\
\midrule
ELC-BERT ACLM-D7 &47.8 & 51.3& 70.0 & 64.8\\
ELC-BERT ACLM-D32 &51.1 &50.7 & 69.8 & 65.7\\
ELC-BERT ACLM-D64 & 51.1& 51.1 &71.0 & 64.8\\
ELC-BERT ACLM-D128 & 50.0 &51.8 &72.1 & 63.5 \\
\bottomrule
\end{tabular}
%}
\caption{Average accuracy scores across the BabyLM evaluation task set for the official baselines, our "plain" ELC-BERT runs, and our ACLM runs over ELC-BERT. We also include the original \citet{charpentier-samuel-2023-layers} result. \label{tab:bblm2024}}
\end{table*}

\section{Conclusions and future work}
For grammatically fined-grained inference tasks, our BLiMP results show that we underperform all models including the baseline, even without ACLM, which we would expect to be similar to the baselines or the original ELC-BERT.  We can straightforwardly suggest that our ELC-BERT attempts were limited by the fact that we trained with a much smaller batch size, although the actual effect of batch size probably needs significantly more exploration, especially why BLiMP specifically is affected by the batch size issue. 

The batch size difference seemed to have a major effect on EWOK and no effect on GLUE for non-ACLM models.  The simplest explanation is that ELC-BERT is simply very sensitive to hyperparameters. To investigate this further, we plan to conduct a hyperparameter study, in particular considering that some of the differences between models are rather small. For example, tuning the learning rate to batch size could be an avenue for optimization, though this has yet to be explored. However, we can contextualize the batch size effect in terms of the performance of our ACLM training regimen.

Our ACLM models were trained under conditions similar to our ELC-BERT runs. Consequently, we did not expect them to actually exceed the LTG-BERT baseline on BLiMP.  We found this to be true, again possibly reflecting the batch size dependence of the task.  But we saw consistent improvements on EWOK and GLUE over both our ELC-BERT-only runs and the LTG-BERT baseline.  These improvement were independent of the dimensionality of the surprisal space, but, in hindsight, this is unsurprising because the input length was already uniform.  

In our entry from last year, we found that reversing the surprisal criterion (effectively choosing the least surprising candidates from the pool) caused a significant delay in result convergence, suggesting that this criterion has an effect, even if we did not have the right conditions to cause it to exceed a baseline BERT model in performance.  We find yet again tantalizing evidence that there are conditions under which controlling the order of learning matters.  EWOK is a world-knowledge-oriented dataset.  We speculate that our learner-directed process may approximate an order that reflects cognitive dependencies among the human tasks---that is, the learner "fine-tunes" successively on increasingly "complex" tasks.  Exploring this requires direct inspection of what learning order is actually chosen by ACLM and empirical investigation in to whether these orders might reflect developmental needs.  

A similar explanation may apply to the consistently higher performance of our ACLM runs on GLUE.  GLUE contains common-sense reasoning entailments, and this may reflect an implicitly preferable learning order that our surprisal criterion is finding.

We emerge from this task optimistic about ACLM as a way of exploring learner-directed strategies for simulating language acquisition through training large language models. There is still a huge methodological space to explore as well as many potentially relevant hyperparameters.  For efficiency and comparability reasons, we adopted ELC-BERT's sentence-independent uniform input length, which likely nullified the effect of varying surprisal space dimensions. However, we believe that sentence length ought to have an effect on the learner's choices in what to focus on next.  In the case of varying sentence length, the method of reduction to a uniform space would likely therefore matter and be an appropriate target of future work.  

We have also focused on surprisal as the measure that steers the interactive learner, but we find it unlikely that a single measure would represent the totality of optimal behaviours.  Therefore, another direction for future work would be testing other measures or combinations thereof.

\section*{Limitations}
Our work is limited to tasks based in English.  We do not have a full analysis of the statistical significances of the differences in the scores.  There are significant areas of the model design and hyperparameter space that we did not explore.  As we replaced RoBERTa with ELC-BERT for this year's BabyLM task, we lose full comparability with last year's results.

\section*{Acknowledgements}
This research was funded in part by the Swedish Research Council (VR) grant (2014-39) for the Centre for Linguistic Theory and Studies in Probability (CLASP). Xudong Hong was funded by the Konrad Zuse School of Excellence in Learning and Intelligent Systems (ELIZA). We thank our student assistant Mattes Alexander Warning for searching hyperparameters for our models. We also thank the anonymous reviewers for their insightful comments.

\bibliography{custom}
\bibliographystyle{acl_natbib}

\appendix
\section{Pre-training details}\label{sec:appendix}

\begin{table*}[t]
    \centering
    \begin{tabular}{l r}
    \toprule
\textbf{Hyperparameter} &  \textbf{Small (Submitted Model)} \\ 
\midrule
Number of parameters &  24M \\
Number of layers & 12 \\
Hidden size is &  384 \\
FF intermediate size &  1 024 \\
Vocabulary size &   6 144 \\
Attention heads & 6 \\
Hidden dropout & 0.1 \\
Attention dropout & 0.1 \\
Training steps & 31 250 \\
Batch size & 512 \\
Initial Sequence length & 128 \\
Warmup ratio & 1.6\% \\
Initial learning rate & 0.005 \\
Final learning rate & 0.005 \\
Learning rate scheduler &  cosine \\
Weight decay &  0.4 \\
Layer norm $\epsilon$ & 1e-7 \\
Optimizer & LAMB \\
LAMB $\epsilon$ & 1e-6 \\
LAMB $\beta_{1}$ &  0.9 \\
LAMB $\beta_{2}$ &  0.98 \\
Gradient clipping &  2.0 \\
Gradient accumulation & 4  \\
\bottomrule
    \end{tabular}
    \caption{Pre-training hyperparameters for ACLM models trained on the STRICT-SMALL track. Note that they are almost identical to the SMALL ELC-BERT model \cite{charpentier-samuel-2023-layers}, with the exception of the batch size and the gradient accumulation.}
    \label{tab:hyperparameters}
\end{table*}

\end{document}